\documentclass[10pt,twocolumn,letterpaper]{article}

\usepackage[pagenumbers]{cvpr} 

\newlength\mytmplen
\usepackage{tikz}          
\usepackage{caption}       
\usetikzlibrary{spy}       

\renewcommand{\paragraph}[1]{\vspace{.25em}\noindent\textbf{#1}}
\usepackage{cuted}   
\usepackage{graphicx}      
\usepackage{booktabs}      
\usepackage[table]{xcolor} 
\usepackage{colortbl}      
\usepackage{mwe} 

\usepackage{graphicx}
\usepackage{amssymb}
\usepackage{amsmath}
\usepackage{booktabs}
\usepackage{bbm}
\usepackage{times}
\usepackage{epsfig}
\usepackage{tabu}
\usepackage{multirow}
\usepackage{amsfonts}       
\usepackage{xpunctuate}
\usepackage{pifont}
\usepackage{color,colortbl}
\usepackage{tabulary,overpic}
\usepackage{booktabs}       
\usepackage[normalem]{ulem}

\usepackage{amsmath}
\usepackage{cite}

\usepackage{scalerel}
\usepackage[accsupp]{axessibility}

\usepackage{xcolor}
\usepackage{enumitem}
\usepackage{arydshln} 
\usepackage{xspace}
\usepackage{booktabs}
\usepackage{colortbl}
\usepackage{multirow}
\usepackage{makecell}
\usepackage{lipsum} 
\usepackage{textcomp}
\usepackage{gensymb} 

\usepackage{tabularx}

\usepackage{booktabs}
\usepackage{caption}
\usepackage{subcaption}

\providecommand{\methodname}{TrackMAE\xspace}

\usepackage[table]{xcolor} 
\usepackage{pifont}        

\usepackage{subcaption} 
\newcommand{\tablestyle}[2]{\setlength{\tabcolsep}{#1}\renewcommand{\arraystretch}{#2}\centering\footnotesize}

\definecolor{cvprblue}{rgb}{0.21,0.49,0.74}
\usepackage[pagebackref,breaklinks,colorlinks,allcolors=cvprblue]{hyperref}

\title{TrackMAE: Video Representation Learning via Track Mask and Predict}

\author{
{Renaud Vandeghen}$^{*~1,2}$\quad
{Fida Mohammad Thoker}$^{*~2}$\quad
{Marc Van Droogenbroeck}$^{1}$\quad
{Bernard Ghanem}$^{2}$\quad\\
$^1$University of Liège \quad
$^2$KAUST\\
{\small $*~$Equal contribution}
}

\begin{document}
 \maketitle
 \begin{abstract}
Masked video modeling (MVM) has emerged as a simple and scalable self-supervised pretraining paradigm, but only encodes motion information implicitly, limiting the encoding of temporal dynamics in the learned representations. As a result, such models struggle on motion-centric tasks that require fine-grained motion awareness. To address this, we propose \methodname, a simple masked video modeling paradigm that explicitly uses motion information as a reconstruction signal. In \methodname, we use an off-the-shelf point tracker to sparsely track points in the input videos generating motion trajectories. 
Furthermore, we exploit the extracted trajectories to improve the random tube masking with a motion-aware masking strategy.
We enhance video representations learned in both pixel and feature semantic reconstruction space by providing a complementary supervision signal in the form of motion targets. We evaluate on six datasets across diverse downstream settings and find that \methodname consistently outperforms the state-of-the-art video SSL baselines, therefore learning more discriminative and generalizable representations. Code available at \url{https://github.com/rvandeghen/TrackMAE}

\end{abstract}    
 \begin{figure}[ht]
    \centering
    \includegraphics[width=\columnwidth]{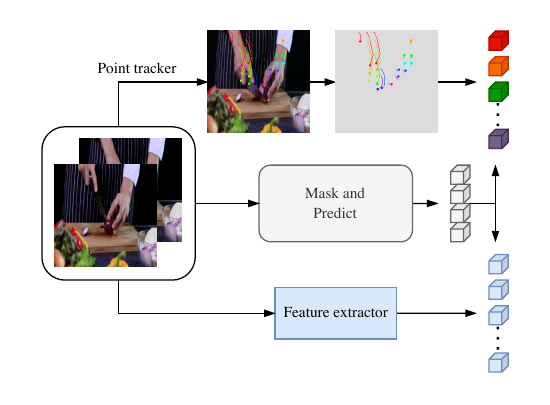}
    \caption{\textbf{\methodname} improves masked video modeling by jointly predicting spatial features and motion trajectories in a mask-and-predict fashion.}
    \label{fig:teaser}
\end{figure}

\section{Introduction}
Self-supervised learning (SSL) has become the default pretraining recipe, allowing models to learn diverse yet powerful representations for different modalities, including text~\citep{Devlin2019BERT, Brown2020Language-arxiv} image~\citep{Chen2020Simple, He2020Momentum, Oquab2024DINOv2, Caron2018Deep, Caron2020Unsupervised, Caron2021Emerging, Bao2022BEiT, Zhou2022Image, He2022Masked} and video~\citep{Tong2022VideoMAE, Wang2023Masked, Huang2023MGMAE} replacing manual labels with pretext objectives that exploit the structure of the raw data. Among video methods, masked video modeling (MVM) stands out for its simplicity and scalability: a high fraction (often 80–95\%) of spatiotemporal tokens is hidden, and a vision transformer is trained to reconstruct them from the visible context~\citep{Tong2022VideoMAE,Sun2023Masked,Huang2023MGMAE,Fan2023MotionGuided}. In practice, videos are decomposed into tubelets (\eg, $2{\times}16{\times}16$) and only visible tokens are encoded, which makes computation efficient and enables training on large corpora. However, the most common instantiation of MVM is pixel reconstruction with random tube masking, which tends to emphasize low-level appearance statistics (color/texture continuity, local smoothness) and under-utilize temporal structure. Real-world videos also exhibit strong temporal redundancy and sparse foreground motion, so masked pixel reconstruction can often be solved via spatial correlations or short-horizon consistency without learning fine-grained dynamics. This mismatch is observed in the downstream performance of such models, as shown in~\citep{Thoker2022HowSevere,Thoker2025SEVERE++-arxiv}. They perform well on appearance-dominated datasets like Kinetics-400~\citep{Kay2017TheKinetics-arxiv} or UCF101~\citep{Soomro2012UCF101-arxiv} but lag on motion-centric tasks such as Something-Something V2~\citep{Goyal2017Something} and FineGym\citep{Shao2020FineGym}, where accurate temporal modeling is crucial.

Recent efforts inject more structure into MVM along two orthogonal axes. One line alters \emph{where} to learn by biasing masks toward informative regions, \eg, selecting high-motion tokens via flow~\citep{Huang2023MGMAE} or motion vectors~\citep{Fan2023MotionGuided}, or adaptively sampling tokens with learned heuristics~\citep{Bandara2023AdaMAE}. Another line changes \emph{what} to learn by replacing pixels with feature-space targets (\eg, HOG, DINO, CLIP), which mitigates pixel-level shortcuts and encourages learning of high-level semantics like object/part structure, attributes, and object–scene relations~\citep{Sun2023Masked,Thoker2025SMILE}. 
While both directions help, they supervise motion only \emph{implicitly}. The model is never asked to predict \emph{how} things move or to maintain identity over time. As a result, improvements in motion sensitivity are limited, especially under high masking ratios where temporal cues must be inferred from few visible tokens. We suggest that temporal correspondence should be a first-hand signal during pretraining, complementing pixel/feature targets rather than competing with them.

In this work, we propose to directly use motion as a training signal by complementing the spatial reconstruction with a motion prediction target. By leveraging the motion prediction produced by a point tracker, our method \textbf{\methodname} (\cref{fig:teaser}) aims to jointly reconstruct both the spatial and motion targets. In particular, we use CoTracker3~\citep{Karaev2025CoTracker3}, a robust and high-quality point tracker, to extract motion trajectories.
\methodname adds two components to masked video modeling: (i) a lightweight \emph{trajectory decoder} that predicts point-track displacements, and (ii) \emph{motion-aware masking} that preferentially samples visible tokens from both high- and low-motion regions. Finally, we show that our motion target complements both pixel- and feature-based spatial targets, leading to on par with or state-of-the-art results on several benchmarks. Our contributions:
\begin{enumerate}
    \item We propose \methodname, a new masked video pretraining scheme with explicit motion awareness based on prediction of tracked point trajectories.
    \item We improve the random tube masking strategy with a motion-aware masking that equally samples visible tokens from high- and low-motion regions.
    \item  Extensive evaluation on multiple datasets demonstrating state-of-the-art performance on motion-centric benchmarks as well as strong generalization capabilities. 
\end{enumerate}

 \section{Related work}
\label{sec:related_work}

\paragraph{Masked video modeling.}
Masked modeling has emerged as a powerful learning paradigm for representation learning for both visual and non-visual modalities like text~\citep{Brown2020Language-arxiv,Devlin2019BERT}, audio~\citep{Huang2022Masked,Chong2023Masked,Gong2022SSAST}, 
images~\citep{Bao2022BEiT, Xie2022SimMIM, He2022Masked, Zhou2022Image, Assran2023SelfSupervised, Gupta2023Siamese, Eymael2024Efficient}, and video~\citep{Tong2022VideoMAE, Huang2023MGMAE, Fan2023MotionGuided, Sun2023Masked, Wang2023VideoMAEV2, Feichtenhofer2022Masked}. The goal is to hide a portion of the input from the model and aim to predict the hidden part from the visible context only.
In vanilla  masked video modeling~\citep{Tong2022VideoMAE}, a considerable portion of the input pixels are typically kept hidden at random, and the model is trained to predict those hidden pixels from the visible portion, thereby encoding useful video representations.  Many follow-up works improve this mask-and-predict task by either leveraging \emph{what to mask}~\citep{Qing2023MAR,Fan2023MotionGuided,Huang2023MGMAE,Bandara2023AdaMAE} or \emph{what to predict}~\citep{Li2023Unmasked,Sun2023Masked,Wei2022Masked,Yang2024MotionMAE,Thoker2025SMILE,Salehi2024SIGMA}. 

In the masking paradigm, MAR~\citep{Qing2023MAR} designed a new masking strategy tailored for action recognition, based on the running cell masking. MGM~\citep{Fan2023MotionGuided} leverages the motion information contained in motion vectors of the raw videos to create a motion-aware masking strategy. MGMAE~\citep{Huang2023MGMAE} follows the same idea of improving the masking based on motion information extracted from an optical flow instead. AdaMAE~\citep{Bandara2023AdaMAE} learns a sampling network that selects regions with high information, which continually improves the masking strategy over training time. 

 Beyond the pixel reconstruction paradigm, MaskFeat~\citep{Wei2022Masked} aims to reconstruct HOG features~\citep{Dalal2005Histograms}. In the same manner, MME~\citep{Sun2023Masked} reconstructs both position changes and the HOG-based shape features. MME builds motion targets based on pre-computed optical flow forming trajectory-like signals, and extracts hand-crafted HOG features around these trajectories for prediction. Such a pipeline requires a heavy pre-processing, camera-motion–sensitive pipeline and produces noisier long-range motion, while we directly extract the trajectories from RGB on the fly. Furthermore, MotionMAE~\citep{Yang2024MotionMAE} aims to reconstruct temporal frame differences, while recent works like SIGMA~\citep{Salehi2024SIGMA} rely on clustered DINO~\citep{Caron2021Emerging} features as the reconstruction target. SMILE~\citep{Thoker2025SMILE} uses CLIP~\citep{Radford2021Learning} features for reconstruction and improves the motion awareness by injecting synthetic motions by copy-pasting segmented objects onto the videos using randomly generated paths. Different from SMILE, our motion information comes from real trajectories from a tracking module representing actual pixel motion.
 
Overall, our approach extracts motion trajectories from a tracking module and leverages them as additional motion reconstruction target and reuses them to enhance the motion awareness in the masking, thereby improving the spatial and motion semantics of the learned video representations.

\paragraph{Learning from motion.} Motion in video data has long served as free supervision in vision, and prior work exploits it for several tasks. For image representation learning, \citet{Wang2015Unsupervised} use visual tracking to create positive pairs, encouraging patches that are linked by a track to have similar features. 
For dense visual correspondence, \citet{Wang2019Learning} learn features by tracking backward and then forward in time (cycle consistency), enabling self-supervised correspondence. \citet{Li2019JointTask} learn dense correspondence by optimizing region tracking and pixel-level matching.

With the emergence of reliable point trackers such as CoTracker3~\citep{Karaev2025CoTracker3}, point tracks have become strong supervisory signals. They can guide attention routing~\citep{Lai2025Tracktention} or supervise time-consistent dense features via clustering~\citep{Salehi2025MoSiC}. Tracktention~\citep{Lai2025Tracktention} injects point-track correspondences into the attention layer of image models, yielding temporally consistent features that handle large motion and turning them into strong video models for depth estimation and colorization.
Similarly, MoSIC~\citep{Salehi2025MoSiC} first clusters long-range tracked trajectories via optimal-transport clustering and then propagates the cluster assignments along tracks to enforce temporal coherence under occlusion and viewpoint change, improving dense representations.
Our approach is inspired by these works to use trajectories for motion injection, albeit with a different objective. Instead of temporal consistency  or label propagation, we aim to improve the motion semantics in masked video modeling representations.

 \section{Methodology}
\label{sec:methodology}

\begin{figure*}[t]
    \centering
    \includegraphics[width=\textwidth]{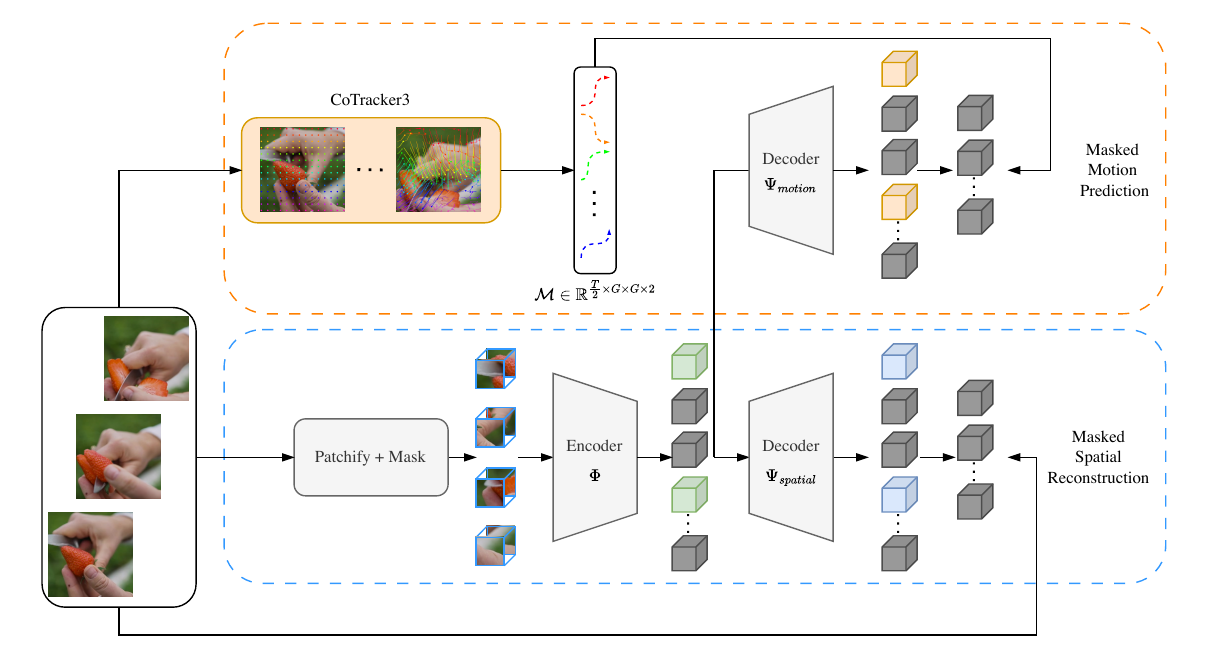}
    \vspace{-1.0cm}
    \caption{\textbf{Overview of \methodname.} In the lower branch, a video clip $\mathbf{V}$ is first patchified and masked. The visible tokens are fed to a ViT encoder $\Phi$. Then the decoder $\Psi_{spatial}$ aims to reconstruct spatial features based on the encoder output. In the upper branch, the input video clip is processed by a CoTracker3 module, extracting sparse point trajectories. The encoder output is then passed to a second decoder $\Psi_{Motion}$, which aims to predict the extracted trajectories. The training objective combines both motion and spatial reconstruction.}
    \label{fig:main_figure}
\end{figure*}
\subsection{Masked Video Modeling}
\label{subsec:masked-video-modeling}

\paragraph{Input.}
Masked video modeling, \eg VideoMAE~\citep{Tong2022VideoMAE}, is an extension of standard image masked modeling MAE~\citep{He2022Masked}. The input is a short clip
$\mathbf{V} \in \mathbb{R}^{T \times H \times W \times 3}$,
which is partitioned into non-overlapping space-time tubelets $\mathcal{C} = \{ \mathbf{c}_i \}_{i=1}^N$ of size $t \times p \times p$, where $t$ is the temporal size of the tubelet and $p$ is the spatial patch size (\eg, $t=2$, $p=16$). This yields $N = \frac{T}{t} \cdot \frac{H}{p} \cdot \frac{W}{p}$ tokens, each corresponding to a local cubic video volume. A tubelet embedding layer (a 3D conv) maps each tubelet in $\mathcal{C}$ into a set of tokens $\mathcal{T} =  \{ \mathbf{\tau}_i \}_{i=1}^N$, $\mathbf{\tau} \in \mathbb{R}^D$, and a fixed positional embedding is added to preserve spatial and temporal order.

\paragraph{Masking.}
Masked video modeling randomly hides a large subset of tokens and asks the model to recover them from the remaining context. A high-ratio masking (typically 90\%) is applied at the token level following a Bernoulli distribution. This produces a visible set $\mathcal{T}^{visible}$ and a masked set of tokens $\mathcal{T}^{masked}$. Such aggressive masking makes reconstruction non-trivial, forcing the model to capture meaningful spatio-temporal dependencies in the video data.

\paragraph{Architecture.}
Masked video modeling relies on a standard ViT-based encoder–decoder architecture. The encoder $\Phi$ takes the visible set of tokens as the input to produce latent representations $\mathbf{Z} = \Phi(\mathcal{T}^{visible})$. The decoder $\Psi$ then maps the encoder's output to produce reconstruction predictions.  To reconstruct the input clip, a complete sequence is formed by inserting learnable \texttt{[MASK]} tokens at the masked positions and adding the same positional embeddings used at input. The decoder $\Psi$  takes this complete sequence and predicts the missing content at every position. 
The goal of the decoder is to predict the space-time tubelets $\hat{\mathcal{C}} = \Psi\left( [\mathbf{Z}, [\texttt{MASK}]] \right)$, of the same size as $\mathcal{C}$.

\paragraph{Reconstruction objectives.}
Since the main goal is to employ a mask-and-predict task, the reconstruction is 
purely performed on masked tokens to prevent any information leakage or shortcut solutions by also predicting the visible tokens.   In most MVM  methods~\citep{Tong2022VideoMAE,Huang2023MGMAE,Fan2023MotionGuided}, the reconstruction is done in the pixel space by directly  optimizing to predict the pixel values of masked space-time tubelets from the set of visible tubelets with the following L2 loss 
\begin{equation} \label{eq:pixel}
	\mathcal{L}_{pixel} = \frac{1}{|\mathcal{T}^{masked}|}  \sum_{i \in \mathcal{T}^{masked} } \left\| \mathbf{c}_i - \hat{\mathbf{c}}_i \right\|_2^2,
\end{equation}
where $ \hat{\mathbf{c}}_i $ is the $i_{th}$ token in the decoder.
To solve this reconstruction task under a strong information dropout, the model has to encode the spatio-temporal dynamics of the input videos, thereby learning useful video representations. 

Beyond the pixel space reconstruction, several works reconstruct in more semantic feature spaces \eg, HOG~\citep{Dalal2005Histograms}, DINO~\citep{Caron2021Emerging, Oquab2024DINOv2}, or CLIP~\citep{Radford2021Learning}. Concretely, video frames are processed to extract per-frame descriptors, which are then aligned to the model’s space–time tokens.
In other words, each space-time pixel tubelet $\mathbf{c}_i$ is projected onto a feature space $\mathcal{F}$, extracting feature tokens $\mathbf{f}_i$ for each tubelet. The model is then optimized to predict the semantic feature values of masked space-time tubelets from the set of visible pixel tubelets, according to the following loss

\begin{equation} \label{eq:feature}
	\mathcal{L}_{feature} = \frac{1}{|\mathcal{T}^{masked}|}  \sum_{i \in \mathcal{T}^{masked} } \left\| \mathbf{f}_i - \hat{\mathbf{f}}_i \right\|_2^2,
\end{equation}
where $ \hat{\mathbf{f}}_i $ is again the $i_{th}$ token in the decoder output $\hat{\mathcal{F}} =  \Psi(\Phi(\mathcal{T}^{visible}),[\texttt{MASK}])$.
Such reconstruction adds an abstraction to the mask-and-predict task, reducing the chances of any shortcuts in pixel reconstruction, and encourages directly modeling high-level video semantics instead of low-level semantics in the pixel space. 

\subsection{TrackMAE: Learning from Motion }
\label{subsec:trackmae}

In this section, we introduce TrackMAE, a masked video pretraining framework that leverages tracked trajectories from CoTracker3~\citep{Karaev2025CoTracker3} in the form of motion prediction and motion-aware masking to enhance temporal awareness in learned video representations.  In particular, we make two additions to the vanilla masked modeling paradigm. First, we integrate motion prediction, \ie predicting a sparse set of trajectories as an additional self-supervision task along with the spatial reconstruction.
Second, we replace the random uniform masking with a motion-aware masking that keeps visible tokens from high- and low-motion regions using displacement magnitudes from the extracted trajectories.

\paragraph{Extracting motion targets.}
As mentioned in \cref{subsec:masked-video-modeling}, the input clip $\mathbf{V} \in \mathbb{R}^{T \times H \times W \times 3}$ is patchified into space-time tubelets $\mathcal{C} = \{ \mathbf{c}_i \}_{i=1}^N$ of size $ t \times p \times p$, where $p$ is the spatial patch size.
Our goal is to approximate the motion in the given input clip $\mathbf{V}$ by tracking the center pixels of patches in the first frame in the subsequent frames with an off-the-shelf tracking module like CoTracker3~\citep{Karaev2025CoTracker3}.
To that end, we sample query points from a uniform grid of size $G \times G$ for the first frame, where $ G = \frac{H}{p}$, and predict their 2D positions ($x,y$) in subsequent frames, extracting a set of motion tracks of shape $T \times \frac{H}{p} \times \frac{W}{p}\times 2$.
For efficiency and to match the shape of input video tokens, we feed the tracking module every other frame, producing motion tokens $\mathcal{M} = \{ \mathbf{m}_i \}_{i=1}^N$ of size 2, matching the size of $\mathcal{C}$. Finally, this set of motion tokens $\mathcal{M}$ is used as the reconstruction target. In practice, we predict the displacement of the point tracks instead of absolute trajectory values. 

\paragraph{Upsampling motion targets.}
By sparsely tracking only one point per $16 \times 16$ patch, we may not accurately capture fine motion displacement in the extracted motion tokens. Ideally, we would like to track as many points per patch as possible to generate denser motion targets. However, dense tracking is expensive, with computational cost proportional to the query grid size $G$. To overcome this issue, we introduce a simple yet effective upsampling trick. Assuming that nearby pixels in a patch behave similarly in terms of motion, we can spatially interpolate extracted sparse motion tokens to simulate denser trajectories per patch. We can spatially upsample the sparse motion tokens $\mathcal{M}$ into a dense set of motion tokens $\mathcal{U}$ of size $\frac{T}{2} \cdot \frac{\upsilon H}{p} \cdot \frac{\upsilon W}{p}$, where $\upsilon$ is the upsampling factor.
This is equivalent to tracking $v^2$ points per patch, generating denser motion targets for reconstruction. We show, in \cref{sec:ablation}, that upsampled targets achieve better downstream performance without any added cost.

\paragraph{Architecture.}
In \methodname, our goal is to jointly solve two prediction tasks in a mask-and-predict manner; one is spatial reconstruction, and the other is motion prediction.  As in \cref{subsec:masked-video-modeling}, we employ an encoder-decoder framework with a common encoder and two separate decoders.  As before, encoder $\Phi$ takes visible tokens  $\mathcal{T}^{visible}$ as input to produce latent features $\mathbf{Z}$.
The two decoders $\Psi_{spatial}$ and $\Psi_{motion}$ take the encoded features $\mathbf{Z}$ with learnable \texttt{[MASK]} tokens and positional embeddings to predict the spatial and motion tokens, respectively, as $\hat{\mathcal{C}} = \Psi_{spatial}\left( [\mathbf{Z}, [\texttt{MASK}]] \right)$ and
$\hat{\mathcal{M}} = \Psi_{motion}\left( [\mathbf{Z}, [\texttt{MASK}]] \right)$.

\paragraph{Objectives.} 
In addition to the spatial reconstruction objective, we additionally optimize for the motion prediction. We follow the masked reconstruction strategy to only predict motion tokens of the hidden portion on the input as
\begin{equation} \label{eq:traj1}
	\mathcal{L}_{motion} = \frac{1}{|\mathcal{T}^{masked}|}  \sum_{i \in \mathcal{T}^{masked} } \left\| \mathbf{m}_i - \hat{\mathbf{m}}_i \right\|_2^2,
\end{equation}
which supervises the motion decoder on masked positions only. The final objective is a weighted sum  of the two objectives as
\begin{equation} \label{eq:traj2}
	\mathcal{L} = \mathcal{L}_{spatial} + \lambda * \mathcal{L}_{motion},
\end{equation}
where $\mathcal{L}_{spatial}$ is either  
$\mathcal{L}_{pixel}$ (\cref{eq:pixel}) or  $\mathcal{L}_{feature}$ (\cref{eq:feature}).
As shown in the experiments, our proposed motion prediction loss $\mathcal{L}_{motion}$  is complementary to both  $\mathcal{L}_{pixel}$ and feature reconstruction   $\mathcal{L}_{feature}$. The overview of the method is shown in \cref{fig:main_figure}.

\begin{figure}[ht]
    \centering
    \includegraphics[width=\columnwidth]{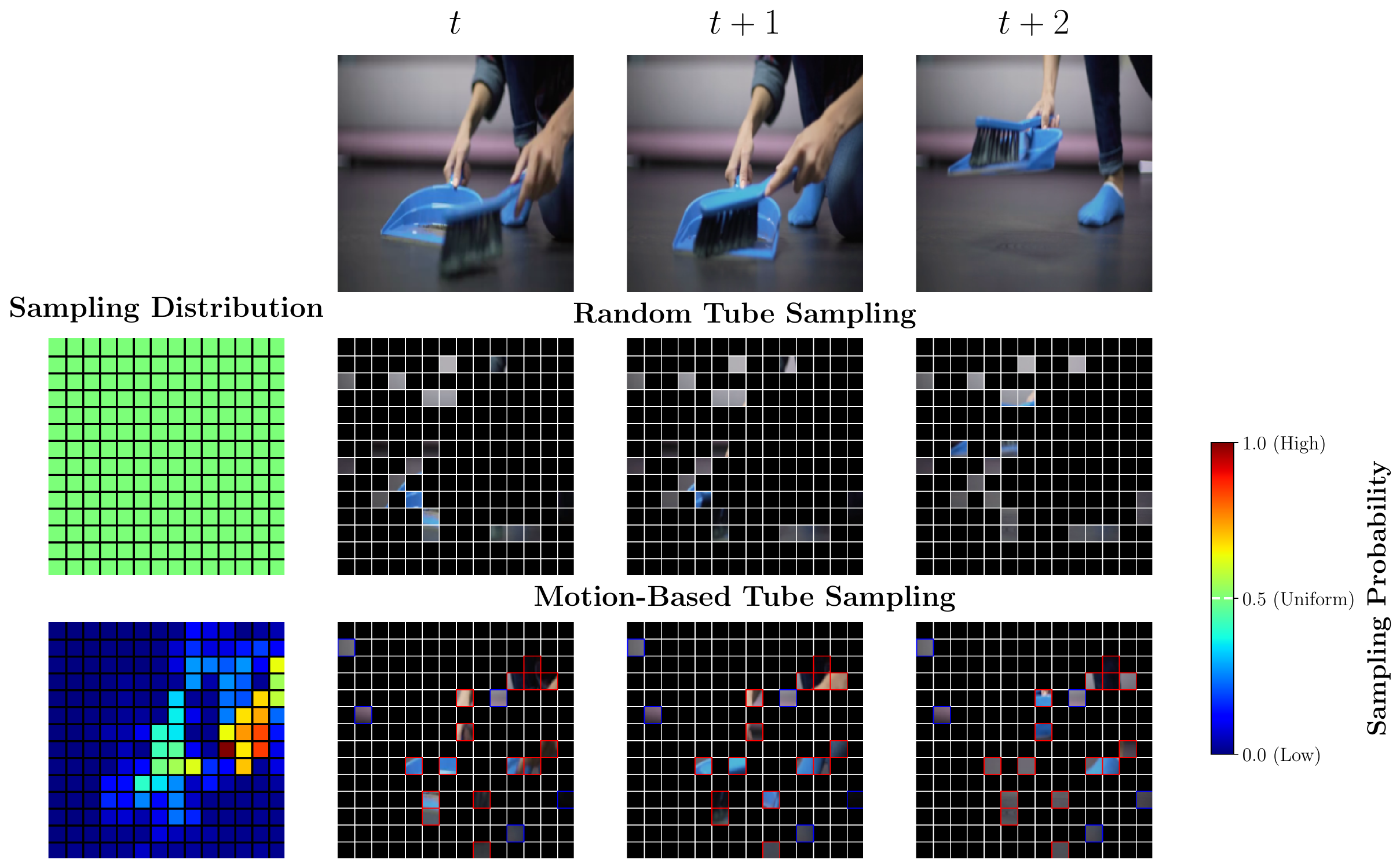}
    \caption{\textbf{Masking comparison.} We show how our motion-based tube masking compares to random tube masking. By explicitly sampling visible tokens, our motion-based sampling distribution ensures that visible tokens cover both motion and static regions. In the motion-based tube sampling, \textbf{{\color{red}red}} squares are sampled from the high-motion and \textbf{{\color{blue}blue}} squares from the low-motion bin.}
    \label{fig:maskingComparison}
\end{figure}

\paragraph{Motion-aware masking.}
The vanilla tube masking used in VideoMAE does not assume any information of motion, leading to random masking maps.
Since the tracker module already gives us motion prediction for trajectory reconstruction, we can also leverage that information to create a motion-guided masking map. In particular, based on the full trajectory predictions, we compute the average displacement $\bar{\mathbf{M}}$, of size $p \times p$, for each query point over the temporal dimension. We then use $\bar{\mathbf{M}}$ as a sampling distribution for the visible tokens. Such distribution can be seen in the first column of the last row in \cref{fig:maskingComparison}. To sample visible tokens from that distribution, we first create 2 uniform bins, containing high- and low-motion samples. We uniformly sample visible tokens from each bin, using a motion ratio $\rho_{motion}$ to control the number drawn per bin.
This masking formulation allows to control where the masking should be done and in what proportion. Such a masking example is depicted in \cref{fig:maskingComparison}, as well as the random tube masking, which can be expressed as a special case of our sampling distribution, using a uniform distribution instead of $\bar{\mathbf{M}}$.
 \section{Experiments}
\label{sec:experiments}
In \cref{sec:implementation}, we describe the pretraining setup that we use in our experiments. In \cref{sec:sota_comparison}, we compare our method against prior methods in different linear probing and full finetuning setups. We run some ablation studies in \cref{sec:ablation} and assess the generalization performance in \cref{sec:severe}. 

\subsection{Implementation Details}
\label{sec:implementation}
Following previous methods, we pretrain a video-based ViT-B model on Kinetics-400 (K400)~\citep{Kay2017TheKinetics-arxiv}. We replace the original tube masking with our motion-guided tube masking, using a motion ratio of $\rho_{motion}=50\%$. We use the default offline CoTracker3 module with a grid size of $14 \times 14$, and use upsampling with $\upsilon{=}2$. For pixel reconstruction, we balance the losses equally with $\lambda{=}1$, and for feature reconstruction, we set $\lambda{=}0.25$. We use a CLIP ViT-B model to extract feature reconstruction targets. Unless stated otherwise, we follow the same hyperparameters as in \citep{Tong2022VideoMAE} and pretrain our model for 800 epochs. More details are in the supplementary material.

\begin{table}[ht]
  \centering
  \caption{\textbf{Linear probing comparison.} We evaluate \methodname on different spatial- and motion-centric benchmarks. For both pixel and feature reconstruction, our method is on par with previous methods for spatial-centric tasks. For motion-centric tasks, our method largely improves previous methods. We report the Top-1\% accuracies of ViT-B models pretrained on K400, and highlight in \textbf{bold} best and \underline{underline} second best results.}
  \setlength{\tabcolsep}{0.15em}
  \small
  \begin{tabular}{l c cc cc} 
    \toprule
    \multirow{2}{*}{\textbf{Method}} & \multirow{2}{*}{\textbf{Target}} &
      \multicolumn{2}{c}{\textbf{Spatial-centric}} &
      \multicolumn{2}{c}{\textbf{Motion-centric}} \\
    \cmidrule(lr){3-4} \cmidrule(lr){5-6}
    & & K400 & HMDB & SSv2 & GYM \\
    \midrule
    \rowcolor{lightgray}
    \textbf{\textit{Pixel Reconstruction}} & & & & & \\
    \hline
    VideoMAE \citep{Tong2022VideoMAE} & Pixel & 20.7 & 37.7 & 17.5 & 23.9 \\
    MVD \citep{Wang2023Masked} & Pixel & 18.7 & 28.6 & 12.2 & 22.7 \\
    MGMAE \citep{Huang2023MGMAE} & Pixel & 24.9 & 41.3 & 16.8 & 26.1 \\
    EVEREST \citep{Hwang2024EVEREST} & Pixel & 14.1 & 30.3 & 14.5 & 23.3 \\
    MGM \citep{Fan2023MotionGuided} & Pixel & 19.8 & 40.3 & 21.7 & 25.8 \\
    \methodname (ours) & Pixel & 25.7 & 40.6 & 23.6 & 29.0 \\
    \hline
    \rowcolor{lightgray}
    \textbf{\textit{Feature Reconstruction}} & & & & & \\
    \hline
    MME \citep{Sun2023Masked} & HOG & 19.1 & 37.1 & 16.6 & 29.0 \\
    SIGMA~\citep{Salehi2024SIGMA} & DINO & 47.5 & 52.3 & 20.8 & 30.1 \\
    SMILE~\citep{Thoker2025SMILE} & CLIP & \textbf{56.2} & \textbf{53.4} & \underline{23.7} & \underline{30.2} \\
    \methodname (ours) & CLIP & \underline{55.2} & \underline{53.1} & \textbf{27.3} & \textbf{31.8} \\
    \bottomrule
  \end{tabular}
  \label{tab:linear-probing}
\end{table}

\subsection{Comparison with State-of-the-Art}
\label{sec:sota_comparison}
\paragraph{Linear Probing.} 
To evaluate the quality of learned video representations, we conduct linear probing experiments across four standard action recognition benchmarks. In this setup, the pretrained encoder is frozen, and a linear classifier is trained on top of it. This isolates the effect of pretraining and removes the influence of fine-tuning dynamics, providing a proper measure of representation quality. We compare \methodname with prior methods based on both pixel and feature reconstruction. For a fair comparison with prior methods, we evaluate using their publicly released ViT‑B checkpoints pretrained on K400. All downstream results are obtained under a unified evaluation protocol. Details of datasets and evaluation are in the supplementary material.

\Cref{tab:linear-probing} presents the linear probing results across multiple benchmarks. In the pixel reconstruction setting, \methodname consistently outperforms the VideoMAE baseline by a margin of  $\sim$5\% across all datasets. This affirms our hypothesis that predicting motion provides a strong self-supervisory signal for learning discriminative video representations. Moreover, \methodname surpasses other motion-aware pixel reconstruction methods such as MGM~\citep{Fan2023MotionGuided} and MGMAE~\citep{Huang2023MGMAE}, particularly on motion-centric datasets like SSv2 and FineGym. This performance gap underscores \methodname’s superior ability to encode fine-grained temporal dynamics.
In the feature reconstruction setting, \methodname again yields consistent gains and outperforms all prior methods, including the state-of-the-art SMILE for the motion-centric tasks. These results demonstrate that motion prediction is not only effective in isolation but also complements semantically rich targets (\eg, CLIP features), enhancing representation learning beyond the pixel space. In summary, these findings confirm that our approach leads to more temporal awareness in the learned representations.

\begin{table}[t!]
\caption{\textbf{Full finetuning comparison.} We evaluate \methodname finetuned on SSv2 and K400 after pretraining the models on K400, outperforming all prior works. We highlight in \textbf{bold} best and \underline{underline} second best results. $\dagger$ means we pretrained on K700.}
\label{tab:sota}
\centering
\scriptsize
\resizebox{\columnwidth}{!}{%
\begin{tabular}{lcccc}
\toprule
\textbf{Method} & \textbf{Backbone} & \textbf{Targets} & \textbf{SSv2 Top-1} & \textbf{K400 Top-1}\\
\midrule
VideoMAE~\citep{Tong2022VideoMAE} & ViT-B & Pixel & 68.5  & 80.0\\
OmniMAE~\citep{Girdhar2023OmniMAE} & ViT-B & Pixel & 69.0 & 80.8 \\
MGM~\citep{Fan2023MotionGuided} & ViT-B & Pixel & 71.1 & 80.8  \\
MGMAE~\citep{Huang2023MGMAE} & ViT-B & Pixel & 68.9 & 81.2 \\
\methodname (ours) & ViT-B & Pixel & 70.1  & 80.8 \\
\hdashline
MME~\citep{Sun2023Masked} & ViT-B & HOG & 70.5 & 81.5 \\
SIGMA~\citep{Salehi2024SIGMA} & ViT-B & DINO & 71.1 & 81.5 \\
SMILE~\citep{Thoker2025SMILE} & ViT-B & CLIP & \underline{72.1} & \underline{83.1} \\
\methodname (ours) & ViT-B & CLIP & \textbf{72.8} & \textbf{83.9} \\
\midrule
VideoMAE~\citep{Tong2022VideoMAE} & ViT-L & Pixel & 74.0 & 85.2\\
\methodname$^\dagger$ (ours) & ViT-L & CLIP & \textbf{75.7} & \textbf{86.7} \\
\bottomrule
\end{tabular}%
} 
\end{table}

\paragraph{Full Finetuning.}
To fully leverage the learned spatiotemporal representations, we perform end-to-end finetuning of both the pretrained backbone and the classification head on downstream datasets. This setup is crucial for assessing the transferability and task-specific adaptability of representations learned through our trajectory-guided pretraining. In contrast to linear probing, which freezes the encoder, full finetuning enables the model to refine its temporal and semantic understanding based on the target task distribution. We evaluate our method on two representative benchmarks: Kinetics-400 and Something-Something V2 (SSv2), covering both appearance- and motion-focused benchmarks. We use the same evaluation protocol as the prior masked video modeling works~\citep{Tong2022VideoMAE,Fan2023MotionGuided,Huang2023MGMAE}, ensuring a fair comparison. Further details are provided in the supplementary material.

We evaluate \methodname under two standard transfer regimes: in-domain transfer (pretraining and finetuning on the same dataset \ie, K400) and cross-domain transfer (pretraining on K400 and finetuning on SSv2). The results, summarized in \cref{tab:sota}, show that \methodname  consistently achieves strong performance across both settings. As with linear probing, we again observe that our \methodname with pixel reconstruction improves the VideoMAE baseline by +0.8\% for in-domain transfer, and +1.6\% for cross-domain transfer. 
Moreover, \methodname with pixel targets is on par or outperforms other prior methods with pixel targets, OmniMAE (1.1\% on SSv2), MGMAE (1.2\% on SSv2), even in full-finetuning settings.
This again validates that adding our motion prediction task to pixel reconstruction is effective  for learning better transferable video representations.

Furthermore, \methodname with CLIP targets outperforms all other prior methods, achieving state-of-the-art performance in all settings. In particular, \methodname outperforms SMILE, which also uses CLIP targets along with synthetic motion infusion for learning motion-aware representations by 0.7\% and 0.8\% on SSv2 and K400, demonstrating better motion encoding without the need for any synthetic motion priors. Finally, we also evaluate our method against VideoMAE with ViT-L backbone, where \methodname shows good scaling properties.
To summarize, adding motion prediction targets complements the spatial reconstruction targets to enrich the learned video representations for both appearance and motion-focused downstream tasks even in full-finetuning settings. We show a detailed comparison with other pretraining settings in the supplementary material. 

 \subsection{Ablations} \label{sec:ablation}
To assess the contribution of each design component in our \methodname framework, we conduct a series of ablation experiments, as summarized in \cref{tab:ablation_reconstructions,tab:ablation_masking,tab:ablation_grid_size,tab:ablation_motion_ratio}. For computational efficiency, we adopt the ViT-S architecture and pretrain on a subset of Kinetics-400, referred to as $\textrm{K400}_{\textrm{s}}$, which includes $1/3$ of the total training videos. Evaluation is performed on both $\textrm{K400}_{\textrm{s}}$ and $\textrm{SSv2}_{\textrm{s}}$, also a reduced version of SSv2.
By default we use pixel and motion trajectories as reconstruction targets, employ a grid size of $14 \times 14$, use random tube masking, set $\lambda{=}1$, and train for 200 epochs, unless specified otherwise.
\begin{table}[!t]
  \caption{\textbf{Reconstruction targets.} 
 Trajectory reconstruction provides a strong supervisory signal to learn useful video representations and complements both pixels and CLIP reconstruction, consistently improving the downstream results.} \centering
  \captionsetup{skip=4pt} 
  \begingroup
  \small                     
  \renewcommand{\arraystretch}{0.90} 
  \setlength{\tabcolsep}{6pt}        
  \begin{tabular}{@{}lcc@{}} 
    \toprule
    \makecell[c]{Reconstruction\\Targets} & $\textrm{K400}_{\textrm{s}}$ & $\textrm{SSv2}_{\textrm{s}}$ \\
    \midrule
    Trajectory only & 46.5 & 53.1 \\
    \cmidrule(lr){1-3}
    Pixels  & 46.0 & 52.2 \\
    Pixels + Trajectory & \textbf{48.9} & \textbf{55.7} \\
    \cmidrule(lr){1-3}
    CLIP & 52.7  & 57.1  \\
    CLIP + Trajectory & \textbf{55.8} & \textbf{61.1} \\
    \bottomrule
  \end{tabular}
  \endgroup
  \label{tab:ablation_reconstructions}
\end{table}

\noindent\textbf{Impact of reconstruction targets.}
\Cref{tab:ablation_reconstructions} analyzes the impact of the different target reconstructions used.  We first observe that trajectory reconstruction as a standalone task is very effective for learning useful video representations, indicating that it can act as a strong self-supervisory signal. Next, we observe that combining trajectory reconstruction with pixel reconstruction outperforms both trajectory-only and pixel-only reconstruction by $(+2.4\%  \textrm{ and } +2.9\%)$ on $\textrm{K400}_{\textrm{s}}$ and  $(+2.6\%  \textrm{ and } +3.5\%)$ on  $\textrm{SSv2}_{\textrm{s}}$, respectively.
Finally, when combined with more high-level semantic targets like CLIP instead of pixel targets, we observe a significant improvement of $(+2.9\%  \textrm{ and } +4.0\%)$. One of the reasons for this is that trajectory targets are closer to pixel targets in the sense that they represent movement of pixels and might rely on the same semantics for solving the reconstruction tasks.  
On the other hand, CLIP targets are highly semantic and largely encode what is present but not how things move, and trajectory prediction fills that gap with a more complementary reconstruction task. We also show our method works with DINO targets in the supplementary.
In summary, this clearly indicates that exploiting motion trajectories as a complementary target strengthens video representation learning. 

\begin{table}[!t]
  \caption{\textbf{Masking strategy.} Replacing random tube masking with our motion-based masking consistently improves the results for both pixels and pixels + trajectory reconstruction.}
  \centering
  \captionsetup{skip=4pt} 
  \begingroup
  \small                     
  \renewcommand{\arraystretch}{0.90} 
  \setlength{\tabcolsep}{6pt}        
  \begin{tabular}{@{}llcc@{}} 
    \toprule
    Target & Masking & $\textrm{K400}_{\textrm{s}}$ & $\textrm{SSv2}_{\textrm{s}}$ \\
    \midrule
    Pixel & Tube & 46.0 & 52.2 \\
    Pixel & Motion-aware & \textbf{46.6} & \textbf{52.6} \\
    \cmidrule(lr){1-4}
    Pixel + Trajectory & Tube & 48.9 & 55.7 \\
    Pixel + Trajectory & Motion-aware & \textbf{49.4} & \textbf{56.2} \\
    \bottomrule
  \end{tabular}
  \endgroup
  \label{tab:ablation_masking}
\end{table}
\noindent\textbf{Impact of motion-aware masking.}
\Cref{tab:ablation_masking} shows the impact of our proposed motion-aware masking over random tube masking. 
We compare our motion-aware masking against random tube masking under two regimes: pixel-only and pixel+trajectory reconstruction. Motion-aware masking yields consistent gains of roughly +0.5\% at no extra computation, as the same trajectories used for supervision are repurposed to construct the masking prior.

\begin{table}[!t]
\caption{\textbf{Impact of dense tracking.} Increasing the tracker grid size consistently improves the results, but at a higher computational cost. Upsampling the tracker's prediction from $14 \times 14$ to $28 \times 28$ yields better performance without any added cost.}
  \centering
  \captionsetup{skip=4pt} 
  \begingroup
  \small                     
  \renewcommand{\arraystretch}{0.90} 
  \setlength{\tabcolsep}{6pt}        
  \begin{tabular}{@{}llcc@{}} 
    \toprule
    Grid Size & Upsampling & $\textrm{K400}_{\textrm{s}}$ & $\textrm{SSv2}_{\textrm{s}}$ \\
    \midrule
    $14 \times 14$ & None & 48.9 & 55.7 \\
    $28 \times 28$ & None & 49.5 & 56.7 \\
    $56 \times 56$ & None & \textbf{50.0} & \textbf{57.0} \\
    \cmidrule(lr){1-4}
    $14 \times 14$ & $14\rightarrow28$ ($\upsilon{=}2$) & {\textbf{50.6}} & {\textbf{57.6}} \\
    $14 \times 14$ & $14\rightarrow56$ ($\upsilon{=}4$) & 50.4 & 57.4 \\

    \bottomrule
  \end{tabular}
  \endgroup
  \label{tab:ablation_grid_size}
\end{table}
\noindent\textbf{Impact of dense tracking.}
As mentioned in \cref{subsec:trackmae}, we track one point per patch from the first frame and initialize query points accordingly on a coarse grid of size $14 \times 14$ to extract sparse motion trajectories. 
In this ablation, we show the impact of extracting and predicting denser trajectories.  In particular, we initialize query points on denser grids ($28\times28$ and $56\times56$) to extract denser trajectories. The results in \cref{tab:ablation_grid_size} show that moving from a coarse to denser prediction consistently improves the results. 

However, these gains come at a cost, since the computational cost of CoTracker3 is directly proportional to the query grid size. 
\Cref{tab:ablation_grid_size} also shows that upsampling the sparse trajectories to denser trajectories via spatial interpolation from $14\rightarrow28$ ($\upsilon{=}2$) yields the strongest gains on both datasets at no cost ($+1.7\%$ and $+1.9\%$). However, upsampling from $14\rightarrow56$ ($\upsilon{=}4$) does not improve over $14\rightarrow28$ ($\upsilon{=}2$) setting.
One of the reasons could be that spatial interpolation exploits the piecewise-smooth nature of local motion and densifies supervision at token locations, strengthening gradients without introducing tracker noise.

\begin{table}[!t]
  \centering
  \captionsetup{skip=4pt}
  \begin{minipage}[t]{0.48\columnwidth}
  \captionof{table}{\textbf{Motion ratio masking.} Equally sampling (50\%) visible tokens from the high motion and low motion bins yields the best results compared to asymmetric sampling.}
    \centering
    \scriptsize
    \renewcommand{\arraystretch}{0.90}
    \setlength{\tabcolsep}{4pt}
    \begin{tabular}{@{}lcc@{}}
      \toprule
      $\rho_{motion}$ & $\textrm{K400}_{\textrm{m}}$ & $\textrm{SSv2}_{\textrm{m}}$ \\
      \midrule
      25 & 48.7 & 55.4 \\
      50 & \textbf{49.4} & \textbf{56.2} \\
      75 & 49.0 & 55.9 \\
      \bottomrule
    \end{tabular}
    \vspace{3pt}
    \phantomsection 
    \label{tab:ablation_motion_ratio}
  \end{minipage}\hfill%
  \begin{minipage}[t]{0.48\columnwidth}
  \captionof{table}{\textbf{Balancing the losses.} Equally balancing the loss ($\lambda = 1.0$) yields the best results compared to unbalanced setups.}
    \centering
    \scriptsize
    \renewcommand{\arraystretch}{0.90}
    \setlength{\tabcolsep}{4pt}
    \begin{tabular}{@{}lcc@{}}
      \toprule
      $\lambda$ & $\textrm{K400}_{\textrm{s}}$ & $\textrm{SSv2}_{\textrm{s}}$ \\
      \midrule
      0.1 & 47.1 & 54.0  \\
      0.5 & 48.2 & 54.6 \\
      1.0 & \textbf{48.9} & \textbf{55.7} \\
      2.0 & 48.8 & 55.7 \\
      \bottomrule
    \end{tabular}
    \vspace{3pt}
    \phantomsection
    \label{tab:ablation_lambda}
  \end{minipage}
\end{table}
\noindent\textbf{Impact of $\rho_{motion}$.}
\Cref{tab:ablation_motion_ratio} indicates how our motion-guided masking behaves when we vary the proportion of masked tokens sampled from high- and low-motion regions. Sampling too few (25\%) or too many (75\%) motion locations slightly degrades performance compared to  50\% motion ratio, showing that equally biasing the mask towards both motion and static parts is a good balance.

\noindent\textbf{Impact of $\lambda$.}
\Cref{tab:ablation_lambda} shows that using the same weight for the losses yields the best results for pixel targets.
However, for clip reconstruction with upsampled trajectory prediction, we found $\lambda =0.25 $ works best, indicating $\lambda$ is sensitive to the type of spatial targets (see supplementary).
 \begin{table*}[t!]
    \centering
    \caption{\textbf{Comparison on SEVERE generalization benchmark~\citep{Thoker2022HowSevere}.} We compare prior video SSL methods on four generalization factors of SEVERE benchmark spanning a total of eight downstream settings. TrackMAE delivers consistently strong or superior performance across these diverse settings in both pixel and feature reconstruction modes, indicating that trajectory-guided masked pretraining improves robustness and generalization of learned video representations.}
    \setlength{\tabcolsep}{2pt}
    \small
    \begin{tabular}{l ccccc cccc}
    \toprule
    \multirow{2}{*}{\textbf{Method}} & \multicolumn{2}{c}{\textbf{Domain shift}} & \multicolumn{2}{c}{\textbf{Sample efficiency ($10^{3}$) }} & \multicolumn{2}{c}{\textbf{Action granularity}} & \multicolumn{2}{c}{\textbf{Task shift}} \\
    \cmidrule(lr){2-3} \cmidrule(lr){4-5} \cmidrule(lr){6-7} \cmidrule(lr){8-9}
    & SSv2 &Gym99 & UCF & GYM & FX-S1 & UB-S1 & UCF-RC$\downarrow$ & Charades \\
    \midrule
    \rowcolor{lightgray} \textbf{\textit{Pixel Reconstruction}} & & & & & & & & \\
    VideoMAE~\citep{Tong2022VideoMAE}  & 68.6 & 86.6 & 74.6 & 25.9 & 42.8 & 65.3 & {0.172} & 17.8 \\
    MVD~\citep{Wang2023Masked}& {70.0} & 82.5 & 67.1 & 17.5 & 31.3 & 50.5 & 0.184 & 16.1 \\
    MGMAE~\citep{Huang2023MGMAE} & 68.9 & 87.2 & 77.2 & 24.1 & 33.7 & 79.5 & 0.181 & 17.9\\
    MGM~\citep{Fan2023MotionGuided}&{71.1} & {89.1} & {78.4} & {26.4} & {38.6} & {86.9} & \textbf{0.152} & {22.5}\\
    \methodname&{70.3} & {88.7} & {79.8} & {31.0} & {41.6} & {85.5} & {0.162} & {20.8}\\
    \rowcolor{lightgray} \textbf{\textit{Feature Reconstruction}} & & & & & & & &\\
    MME~\citep{Sun2023Masked}&{70.1} & {89.7} & {79.2} & {29.8} & \underline{55.5} & {87.2} & \underline{0.155} & {23.6}\\
    SIGMA~\citep{Salehi2024SIGMA} &{70.9} & {89.7} & \underline{84.1} & {28.0} & {55.1} & {79.9} & {0.169} & {23.1}\\
     SMILE~\citep{Thoker2025SMILE}  &\underline{72.1} & {90.8} & \underline{}{86.4} & \textbf{35.1} & {55.1} & \underline{88.3} & {0.170} & \textbf{32.5}\\
     
     \methodname w/o motion &{71.9} & {90.0} & {85.5} & {31.4} & {55.1} & {74.6} & {0.170} & {27.1}\\
     
     \methodname & \textbf{72.8} & \textbf{91.1} & \textbf{86.7} & \underline{34.4} & \textbf{59.0} & \textbf{90.1} & {0.170} & \underline{30.5}\\
    \bottomrule
    \end{tabular}

    \label{tab:severe-performance}
\vspace{-0.3cm}
\end{table*}

\subsection{Downstream Generalization on SEVERE}
\label{sec:severe}
Next, following~\citep{Salehi2024SIGMA,Thoker2025SMILE}, we assess the robustness of \methodname beyond standard action recognition on the SEVERE benchmark proposed in~\citep{Thoker2022HowSevere} and extended in~\citep{Thoker2025SEVERE++-arxiv}. It consists of a controlled suite of eight evaluations designed to probe four aspects of generalization: \textit{domain shift}, \textit{sample efficiency}, \textit{action granularity}, and \textit{task shift}. Concretely, we measure transfer under distribution shift on Something-Something V2 and FineGym (Gym99), low-data finetuning with 1K examples on UCF101 and FineGym, fine-grained discrimination on FX-S1 and UB-S1 splits of FineGym, and non-standard objectives via temporal repetition counting on UCFRep~\citep{Zhang2020ContextAware} and multi-label classification on Charades~\citep{Sigurdsson2016Hollywood}. All experiments follow the official SEVERE protocols and are reported in \cref{tab:severe-performance}. Benchmark and evaluation details are reported in the supplementary material.

\noindent\textbf{Domain shift.}
On SSv2 and Gym99, \methodname with CLIP+motion targets attains the strongest performance among masked video modeling approaches, slightly improving over SMILE and other feature-based baselines. Importantly, even the pixel-only \methodname variant improves over VideoMAE and remains competitive with motion-aware designs such as MGM and MGMAE, showing that explicit trajectory prediction enhances robustness to distribution shifts even when training purely in pixel space.
\newline
\noindent\textbf{Sample efficiency.}
In the 1K-sample setting (UCF$10^{3}$, Gym$10^{3}$), the pixel-based \methodname outperforms all prior pixel-based methods and even surpasses feature-based methods like MME and SIGMA, indicating that injecting motion structure into masked prediction benefits low-shot recognition without relying on high-level feature targets. 
\newline
\noindent\textbf{Action granularity.}
For fine-grained splits FX-S1 and UB-S1, \methodname with CLIP+motion achieves the best results, and the pixel-based \methodname variant also improves over or is on par with motion-guided baselines like MGM and MGMAE. These gains underline that \methodname is particularly effective for capturing subtle temporal and spatial differences required in fine-grained action classification.
\newline
\noindent\textbf{Task shift.}
For the task shift, \methodname shows improvements over baselines VideoMAE and MGMAE in the pixel space but is on par or slightly worse than the current best method SMILE in the feature reconstruction.
\newline
\noindent\textbf{Summary.} In the pixel space, \methodname outperforms most prior works \eg VideoMAE and MGMAE across all factors by significant margins, showing strong generalization capability. Furthermore, adding our motion prediction targets to CLIP-only reconstruction (denoted as \methodname w/o motion) also improves results, indicating that the generalization capability is enhanced by using motion trajectories.

 \section{Discussions}

\label{sec:discussion}

\noindent\textbf{Computational cost.}
Point tracking comes at a non-negligible cost. In practice, we observe that the pretraining time increases by ~50\%. However, with stronger performance on linear probing and generalization, the tradeoff is reasonable. Furthermore, we are able to mitigate larger cost due to denser tracking with our upsampling strategy enabling our method to extract motion targets on the fly.
\newline
\noindent\textbf{Robustness to Point Tracks.}
Point trackers are prone to incorrect prediction, often due to very high motion or occlusion. To evaluate the robustness of \methodname under jittered trajectories, we trained our model to reconstruct noisy predictions, either spatially or temporally, using Gaussian noise, see supplementary. We observe a performance degradation of around -0.5\%, showing robustness to noisy point tracks.
We also show \methodname is generalizable to different point trackers \eg TAPNext~\citep{Zholus2025TAPNext} in the supplementary.

 \section{Conclusion}
\label{sec:conclusion}
We introduce \methodname, a new masked video modeling paradigm based on motion prediction. In particular, we leverage a sparse point tracker to create motion-based reconstruction signal. 
To mitigate the computational cost needed to obtain denser trajectories, we show that spatially interpolating the tracker output yields better performance without additional cost. 
We also  use the motion trajectories to propose a new motion-aware masking strategy that further improves the downstream performance.
Across linear probing, full fine-tuning, and SEVERE, \methodname shows consistent gains on appearance- and motion-centric tasks, translating into stronger discrimination and broader generalization than prior video SSL methods.
\clearpage
\section*{Acknowledgments}
The present research benefited from computational resources made available on Lucia, the Tier-1 supercomputer of the Walloon Region, infrastructure funded by the Walloon Region under the grant agreement n°1910247. The research reported in this publication was supported by funding from King Abdullah University of Science and Technology (KAUST) - Center of Excellence for Generative AI, under award number 5940. For computing time, this research used Ibex managed by the Supercomputing Core Laboratory at King Abdullah University of Science \& Technology (KAUST) in Thuwal, Saudi Arabia. We acknowledge EuroPC JU for awarding the project ID EHPC-DEV-2025D10-008 access to Leonardo on Leonardo Booster hosted by CINECA, Italy and access to MareNostrum5 on MN5 ACC hosted by BSC, Spain.

\clearpage
\setcounter{page}{1}
\maketitlesupplementary

\section{Detailed SOTA Comparison}

\begin{table*}[t!]
\caption{\textbf{Detailed SOTA comparison of masked video modeling methods on Something-Something V2 and Kinetics-400 for full finetuning action recognition.} Our \methodname outperforms many supervised approaches and achieves the best performance masked video modeling methods with similar pretraining setups. $\dagger$ means we pretrained on K700.}
\centering
\renewcommand{\arraystretch}{1.2}
\begin{tabular}{lllllll}
\toprule
 & \multicolumn{3}{c}{\textbf{}} &\multicolumn{1}{c}{\textbf{SSv2 Pretraining}} & \multicolumn{2}{c}{\textbf{K400 Pretraining}} \\ 
 \cmidrule(r){5-5}  \cmidrule(r){6-7} 
\textbf{Method} & \textbf{Backbone} & \textbf{Targets} & \textbf{Epochs} & \textbf{SSv2 Top-1}  & \textbf{SSv2 Top-1} &\textbf{K400 Top-1}\\
\midrule
\rowcolor{lightgray}  \textit{\textbf{Supervised}} & & & & & & \\
    \hline
Mformer~\citep{Patrick2021Keeping} &Mformer-B &{-} & - & - &66.7& 79.7  \\
VideoSwin~\citep{Liu2022Video} & Swin-B & {-} & -  & - &69.6& 80.6    \\
TimeSformer~\citep{Bertasius2021Is} & ViT-B & {-} & - & - &59.5 &80.7 \\
MViTv1~\citep{Fan2021Multiscale} & MViTv1-B & {-} & - & - &67.7 &80.2  \\
MViTv2~\citep{Li2022MViTv2} & MViTv2-B & {-} & - & - &70.5 &82.9  \\
Uniformer-B~\citep{Li2022UniFormer} &Uformer-B &{-} & - &- &71.2 & 83.0  \\
\hline
\rowcolor{lightgray}  \textit{\textbf{Self-supervised}} & & & & & & \\
VideoMAE~\citep{Tong2022VideoMAE} & ViT-B & Pixel & 800 & 69.6 &  68.5  & 80.0\\
VideoMAE~\citep{Tong2022VideoMAE} & ViT-B & Pixel & 1600 & 69.6 &  -  & 81.5\\

CMAE-V~\citep{Lu2023CMAEV-arxiv} & ViT-B &Pixel & 800 & 69.7 & - & 80.2 \\
CMAE-V~\citep{Lu2023CMAEV-arxiv} & ViT-B &Pixel & 1600 & 70.5 & - & 80.9 \\
OmniMAE~\citep{Girdhar2023OmniMAE} & ViT-B &Pixel &800 &  69.5 & 69.0 & 80.8 \\

MGM~\citep{Fan2023MotionGuided} & ViT-B &Pixel &800 &  70.6  & 71.1 & 80.8  \\
MGM~\citep{Fan2023MotionGuided} & ViT-B &Pixel &1600 &  71.8  & - & 81.7  \\

MGMAE~\citep{Huang2023MGMAE} & ViT-B &Pixel & 800 & 71.0  & 68.9  & 81.2 \\
MGMAE~\citep{Huang2023MGMAE} & ViT-B &Pixel & 1600 & 72.0  & -  & 81.8 \\

MME~\citep{Sun2023Masked} & ViT-B &HOG &800 &  70.0  & 70.5 & 81.5 \\
MME~\citep{Sun2023Masked} & ViT-B &HOG &1600 &  -  & - & 81.8 \\

SIGMA~\citep{Salehi2024SIGMA} & ViT-B &DINO&800 &  71.2  & 71.1  & 81.5 \\

SMILE~\citep{Thoker2025SMILE} & ViT-B &CLIP&600 &  \textbf{72.5}  & 72.1  & 83.1 \\
SMILE~\citep{Thoker2025SMILE} & ViT-B &CLIP&1200 &  - & 72.4  & 83.4 \\

TrackMAE (Ours) & ViT-B &CLIP &800 &  \textbf{72.5}  & \textbf{72.8}  & \textbf{83.9} \\

\midrule
VideoMAE~\citep{Tong2022VideoMAE} & ViT-L & Pixel & 1600 & - & 74.0 & 85.2\\
\methodname$^\dagger$ (ours) & ViT-L & CLIP & 800 & - & \textbf{75.7} & \textbf{86.7} \\
\bottomrule
\end{tabular}
\label{tab:sota_supp}
\end{table*}

In the main paper, we restricted comparisons to self-supervised masked video models with the same backbone and similar pretraining schedule. We now broaden that comparison to include both self-supervised (MVM variants with pixels/HOG/DINO/CLIP targets) and supervised baselines on K400 and SSv2, reported with the ViT-B backbone and a range of pretraining epochs. We evaluate in the full finetuning setup on K400 and SSv2 datasets to assess in-domain and cross-domain transfer. We also evaluate the data and model scaling behaviors of \methodname. The results are shown in~\cref{tab:sota_supp}.

\paragraph{Kinetics-400 (in-domain pretraining and finetuning).} On K400, TrackMAE attains 83.9 Top-1 with 800 epochs, outperforming all listed masked-video baselines trained for equal or longer schedules: VideoMAE~\citep{Tong2022VideoMAE} (80.0/81.5 at 800/1600), CMAE-V~\citep{Lu2023CMAEV-arxiv} (80.2/80.9), OmniMAE (80.8), MGM (80.8/81.7), MGMAE~\citep{Huang2023MGMAE} (81.2/81.8), MME~\citep{Sun2023Masked} (81.5/81.8), and SIGMA~\citep{Salehi2024SIGMA} (81.5). TrackMAE also edges SMILE~\citep{Thoker2025SMILE} (83.1 at 600 and 83.4 at 1200), indicating a stronger accuracy–compute trade-off at shorter schedules. Compared to supervised architectures, TrackMAE surpasses MViTv2-B~\citep{Li2022MViTv2} (82.9) and Uniformer-B~\citep{Li2022UniFormer} (83.0), despite using the standard ViT-B backbone and a self-supervised objective.

\paragraph{SSv2 (cross-domain: K400→SSv2).} When pretrained on K400 and finetuned on SSv2, TrackMAE reaches 72.8, improving over SMILE~\citep{Thoker2025SMILE} (72.1 at 600; 72.4 at 1200) and clearly ahead of VideoMAE~\citep{Tong2022VideoMAE} (68.5), SIGMA~\citep{Salehi2024SIGMA} (71.1), MGM~\citep{Fan2023MotionGuided} (71.1), and MGMAE~\citep{Huang2023MGMAE} (68.9). It also exceeds the supervised counterparts reported under the same column (e.g., VideoSwin~\citep{Liu2022Video} 69.6, MViTv2-B~\citep{Li2022MViTv2} 70.5, Uniformer-B~\citep{Li2022UniFormer} 71.2). These results indicate better domain transfer to motion-centric SSv2, consistent with the hypothesis that explicit motion supervision complements CLIP-space semantics.

\paragraph{SSv2 (in-domain: SSv2→SSv2).} With SSv2 pretraining, TrackMAE attains 72.5, matching the best reported MVM result in the table (SMILE~\citep{Thoker2025SMILE} at 72.5) and exceeding other MVM baselines like MGMAE~\citep{Huang2023MGMAE} (72.0 at 1600), MGM~\citep{Fan2023MotionGuided} (71.8 at 1600), CMAE-V~\citep{Lu2023CMAEV-arxiv} (70.5 at 1600), VideoMAE~\citep{Tong2022VideoMAE} (69.6 at 800), and OmniMAE~\citep{Girdhar2023OmniMAE} (69.5). This suggests TrackMAE’s motion signal remains beneficial even when the pretraining domain already contains substantial temporal variation.

\paragraph{Method scaling.} We scaled \methodname for both the dataset and model sizes. We pretrained our model on K700, with a ViT-L backbone, using ViT-L CLIP features as spatial targets. We report an improvement of $+1.7\%$ and $+1.5\%$ over VideoMAE with the same backbone, and $+2.9\%$ and $+3.1\%$ over our ViT-B model, showing that our model is able to scale well on larger dataset and model sizes.

In summary, TrackMAE achieves state-of-the-art results among masked-video models under comparable settings and is competitive with, or outperforms, strong supervised models. The pattern supports the central claim that explicit motion supervision paired with CLIP-space reconstruction produces video representations that are both semantically strong and motion-aware, yielding robust transfer in-domain and under-domain shift.

\begin{table*}[htbp]
\caption{\textbf{Additional ablations on  our proposed \methodname.}
The default setting uses pixel reconstruction and motion prediction with a grid size of 14, masking ratio of 90\%, $\lambda=1$, and two separate decoders.}
\label{tab:additional}
    \centering
    \tablestyle{4 pt}{1.12}
    \begin{subtable}{.30\linewidth}
        \centering
        \begin{tabular}{llcc}
        \toprule
        Ratio & $\textrm{K400}_{\textrm{s}}$ & $\textrm{SSv2}_{\textrm{s}}$ \\
        \midrule
         95\% &   48.1 & 54.4 \\
         90\% &   \textbf{49.4} & \textbf{56.2} \\
         80\% &   {49.2} & {56.0} \\
        \bottomrule
        \end{tabular}
        \caption{\textbf{Masking ratio.} Masking 90\% of the tokens, as in previous methods, works best.
        }
        \label{tab:supp_masking_ratio}
    \end{subtable}  
    \hspace{0.2cm} 
    \begin{subtable}{.30\linewidth}
        \centering
        \begin{tabular}{lcccc}
        \toprule
        $\lambda$ & GridSize & $\textrm{K400}_{\textrm{s}}$ & $\textrm{SSv2}_{\textrm{s}}$ \\
        \midrule
        0.25 & 28 & 54.4 &    59.6 \\
        0.5 &  28 & 55.1 &    60.1 \\
        1.0 &  28 & 55.8 &    61.1 \\
        \hline
        0.25 & $14\rightarrow 28$ &  \textbf{55.9} & \textbf{61.1} \\
        0.50 & $14\rightarrow 28$  & 55.0 & 60.4 \\
        1.0 &  $14\rightarrow 28$   &  54.3 &    59.7 \\
        \bottomrule
        \end{tabular}
        \caption{\textbf{Sensitivity of $\lambda$}. For CLIP reconstruction and upsampled trajectory prediction, $\lambda =0.25 $ achieves the best performance.
        }
        \label{tab:supp_lambda}
    \end{subtable}%
    \hspace{0.2cm} 
    \begin{subtable}{.30\linewidth}
        \centering
        \begin{tabular}{llcc}
        \toprule
        Noise & $\textrm{K400}_{\textrm{s}}$ & $\textrm{SSv2}_{\textrm{s}}$ \\
        \midrule
         None &   \textbf{49.4} & \textbf{56.2} \\
         Spatial &   {48.9} & {55.7} \\
         Temporal &   {48.8} & {55.8} \\
        \bottomrule
        \end{tabular}
        \caption{\textbf{Impact of noisy tracks.} Adding spatial or temporal noise to the tracker trajectories does not affect our method much. 
        }
        \label{tab:supp_noise}
    \end{subtable}  

    \begin{subtable}{.30\linewidth}
        \centering
        \begin{tabular}{lcc}
        \toprule
        \textbf{Method} & K400$_s$ & SSv2$_s$ \\
        \midrule
        TrackMAE-DINO & 55.2 & 60.6 \\
        TrackMAE-CLIP & \textbf{55.8} & \textbf{61.1} \\
        \bottomrule
        \end{tabular}
        \caption{\textbf{Target type comparison}. Reconstructing DINO or CLIP features works well for spatial reconstruction signal.}
        \label{tab:supp_target}
    \end{subtable}
    \hspace{0.2cm}
    \begin{subtable}{.30\linewidth}
        \centering
        \begin{tabular}{lcc}
        \toprule
        \textbf{Method} & K400$_s$ & SSv2$_s$ \\
        \midrule
        TAPNext & 55.3 & 60.7 \\
        CoTracker3 & \textbf{55.8} & \textbf{61.1} \\
        \bottomrule
        \end{tabular}
        \caption{\textbf{Tracker comparison}. Leveraging different off-the-shelf point tracker leads to overall close performance.}
        \label{tab:supp_tracker}
    \end{subtable}
    \begin{subtable}{.30\linewidth}
        \centering
        \begin{tabular}{lccc}
        \toprule
        Decoder &   $\textrm{K400}_{\textrm{s}}$ & $\textrm{SSv2}_{\textrm{s}}$ \\
        \midrule
        Joint &   48.7 &    55.5 \\
        Separate & \textbf{49.4} & \textbf{56.2} \\
        \bottomrule
        \end{tabular}
        \caption{\textbf{\textit{Joint} or \textit{Separate} decoders}. Using separate decoder for motion prediction outperforms using the joint decoder.
        }
        \label{tab:supp_decoder}
    \end{subtable}%
\end{table*}

\section{Additional Ablations}

\paragraph{Impact of masking.} \Cref{tab:supp_masking_ratio} shows the impact of the masking ratio used during pretraining on the finetuning performance. We observe that a masking ratio of 90\%, similar to previous methods~\citep{Tong2022VideoMAE, Huang2023MGMAE} works best for the pixel reconstruction. However, we found that for clip reconstruction, a masking ratio of 80\% works better, also observed in~SMILE~\citep{Thoker2025SMILE}.

\paragraph{Impact of loss balancing.} We analyze the effect of $\lambda$ with respect to the grid size for CLIP feature reconstruction. The results are shown in \cref{tab:supp_lambda}. Without upsampling, the best results are achieved with $\lambda=1.0$. However, when we upsample from $14\rightarrow28$, the best value is obtained with $\lambda=0.25$. In practice, we favor the upsampling version for efficiency reasons, as already mentioned in the main paper.

\paragraph{Impact of noisy tracks.} As mentioned in the main paper, our method shows robustness against noisy target tracks. The results are shown in \cref{tab:supp_noise}. We observe that our method is robust for both spatial and temporal noise, meaning that even under a noisy tracker, our model would still be able to learn meaningful features.

\paragraph{Impact of feature targets.} \Cref{tab:supp_target} highlights how our method behaves when reconstructing DINO features~\citep{Caron2021Emerging} instead of CLIP features. CLIP features work best, but results obtained with DINO features show seamless integration of other type of features.

\paragraph{Impact of point trackers.} Similar to \cref{tab:supp_noise}, we analyze in \cref{tab:supp_tracker} the influence of the point tracker used for trajectory generation. We observe that TrackMAE is generalizable to different types of point trackers, as TAPNext~\citep{Zholus2025TAPNext} shows reasonable performance without any tweaking. This demonstrates the usability of the future point trackers as a plugin in the TrackMAE framework. 

\paragraph{Impact of separate decoding.} We study in \cref{tab:supp_decoder} the impact of training with a joint or separate decoders to reconstruct both spatial and trajectory targets. As shown in the table, separate decoders work best, which may indicate that involving the same decoder for both tasks may lead to information leakage, degrading the signal.

\begin{figure}[ht]
    \centering
    \includegraphics[width=\columnwidth]{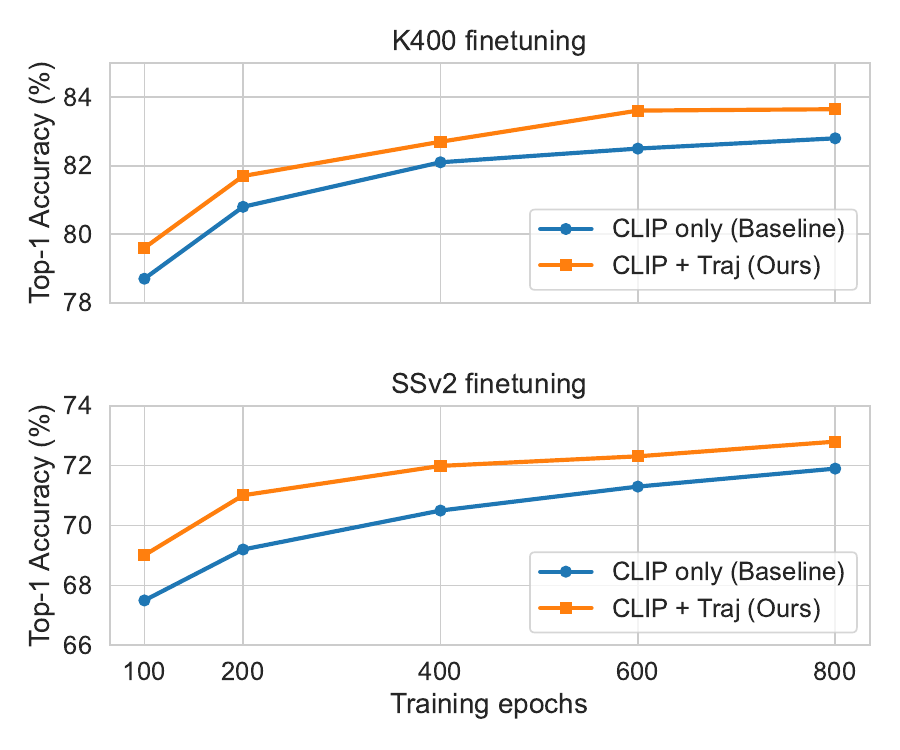}
    \caption{\textbf{Training evolution.} We report the Top-1 Accuracy for K400 and SSv2 finetuning at different pretraining epochs. Our model trained with both CLIP and trajectory reconstructions consistently outperforms the CLIP reconstruction only.}
    \label{fig:supp_training_evolution}
\end{figure}
\section{Convergence Results}
We analyze the convergence behavior of our approach compared to the baseline.
We evaluate our CLIP + Trajectory and CLIP-only reconstruction at different pretraining epochs on both K400 and SSv2 finetuning tasks. ~\Cref{fig:supp_training_evolution} shows that our model trained with both the spatial and trajectory targets consistently outperforms the model trained with the spatial targets only. We observe that with fewer epochs, our performance is much better than the baseline; however, the gap is narrowed with more pretraining epochs. 

\section{Discussion on Motion Masking}
\label{sec:supp_other_masking}

In this section, we expand the discussion on our motion-aware strategy.  In particular, we expand the discussion on different sampling distributions in~\cref{sec:supp_sampling_distribution}, and the different sampling strategies in~\cref{sec:supp_sampling_strategy} based on  motion trajectories.

\begin{figure}[ht]
    \centering
    \includegraphics[width=\columnwidth]{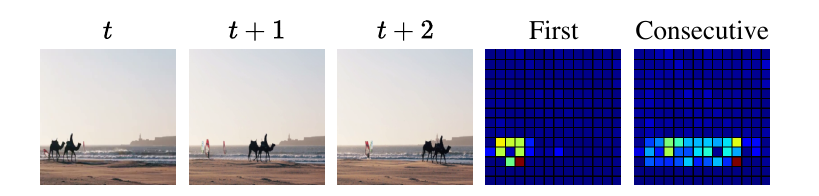}
    \caption{\textbf{Sampling strategies.} We show how we can use the motion information to create different sampling distribution.}
    \label{fig:supp_sampling_strategy}
\end{figure}
\subsection{Sampling Distribution in Masking}
\label{sec:supp_sampling_distribution}
As mentioned in the main paper, we use the motion trajectories to create a sampling distribution.
We explore two different ways to do this: (1) we accumulate the displacement made by each point of the grid through time with respect to the first frame, (2) we accumulate the displacement made by each point of the grid through time in the next consecutive frame. In other words, the first strategy gives the average motion information over time of \textit{how much} a given point is moving, without taking into account the trajectory. The second strategy gives the average motion information over time of \textit{how much} and \textit{where} a given point is moving. Those 2 sampling strategies are depicted in \cref{fig:supp_sampling_strategy}.  

Both sampling strategies have their own merits, \ie the first strategy will put some emphasis towards tokens that are likely to move, without guarantee that they are seen in the following frames. Our intuition is that it forces the model towards learning some motion dynamics of the different moving objects. For the second strategy, our intuition follows the same reasoning, but with visible tokens sampled along the trajectory. In practice, we see both version works on par, with slightly better results for the first strategy, which is used for the results in the main paper. It is also worth mentioning that whatever the strategy used, it is always impacted by the input data. We observe that the K400 dataset is prone to have erratic movements, which may lead to poor motion information. In such a case, our sampling distribution tends to behave more like a uniform sampling, thus falling back to the original random tube masking strategy.

\begin{figure}[ht]
    \centering
    \includegraphics[width=\columnwidth]{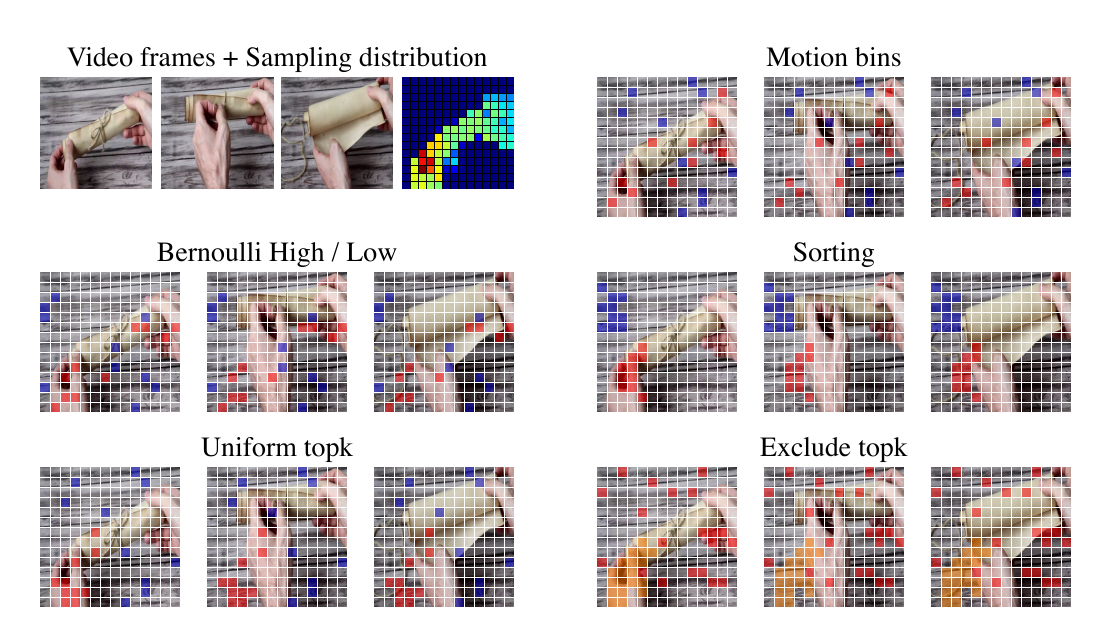}
    \caption{\textbf{Masking strategies.} We show how we can use our sampling distribution to create different masking maps. {\color{red}{Red}} squares are high-motion visible tubes, {\color{blue}{blue}} squares are low-motion visible tokens.  
For the "exclude topk" strategy, we also show {\color{orange}{orange}} the tokens excluded from being sampled. 
    }
    \label{fig:supp_masking_strategy}
\end{figure}
\subsection{Sampling Strategy in Masking}
\label{sec:supp_sampling_strategy}

Besides different sampling strategy, there are also many different ways to sample visible tokens from it. For computational reasons and for its intuitive soundness, we only evaluated the motion bins strategy described in the main paper. However, we describe below some other strategies.

\paragraph{Motion bins.} We sample visible tokens from high- and low-motion regions of the sampling distribution using 2 uniform bins. Depending on the value of $\rho_{motion}$, we control the number of tokens sampled coming from each bins. Our intuition is that this strategy gives, on average, a good balance between high- and low-motion visible and masked tokens. This is the strategy used in the main paper.

\paragraph{Bernoulli High/Low.} Instead of uniformly sampling from high- and low-motion regions, we can sample using a Bernoulli distribution. For the high-motion tokens, we directly sample from the sampling distribution, and for the low-motion token, we sample from the “1-motion” distribution, excluding tokens already sampled from the high-motion part. Similarly as for the motion bins strategy, we can use the same parameter to control the number of tokens for each high- and low-motion regions. However, we visually observe that this strategy tends to create blob regions, which would usually hurts training.

\paragraph{Sorting.} Based on the previous strategy, we can directly sort high- and low-motion tokens and follow their ordering, instead of sampling from the distribution. This strategy creates bigger blobs, which we believe would hurt even more the training.

\paragraph{Uniform topk.} To find a better balance between Bernoulli and sorting, we can uniformly sample from the Top-k most moving tokens. However, in order to sample from a smaller set than in the motion bins strategy, we need to specify how many samples are used in the Top-k operation, adding a new hyperparameter to tune.

\paragraph{Exclude topk.} In contrast to previous strategies, which use the motion to guide the sampling of visible tokens, we can use the motion information to decide where \textit{we should not} sample visible tokens. From the Top-k most moving tokens, we can chose to explicitly remove them from the sampling distribution, as learning to reconstruct those tokens may be interesting. Then, we can use any strategy to sample the visible tokens. All sampling strategies are shown in \cref{fig:supp_masking_strategy}

We leave the exploration of such masking strategies, their impact on different pretraining data and downstream tasks for future work.

\section{Experimental Details}
\label{sec:exp_details}

\subsection{Datasets}

\paragraph{Kinetics-400~\citep{Kay2017TheKinetics-arxiv}}  (K400) is a large-scale YouTube-sourced corpus with 400 human action categories and over 306k short clips. It remains a canonical benchmark for learning generalizable video representations.

\paragraph{Something-Something V2~\citep{Goyal2017Something}}  (SSv2) emphasizes object-centric, first-person interactions that differ markedly from K400’s Internet footage. It comprises 168{,}913 training and 24{,}777 test samples spanning 174 categories, stressing temporal reasoning and commonsense dynamics.

\paragraph{UCF-101~\citep{Soomro2012UCF101-arxiv}} (UCF) collects 9{,}537 training and 3{,}783 test clips from YouTube across 101 action classes. Although coarser in granularity and overlapping with K400 categories, it is widely used for transfer evaluation in self-supervised video learning.

\paragraph{HMDB-51~\citep{Kuehne2011HMDB}} (HMDB) contains 6{,}766 clips from various sources (films, archives, web videos) covering 51 classes (at least 100 videos per class). Its heterogeneity challenges models to cope with diverse cinematic and real-world content.

\paragraph{FineGYM~\citep{Shao2020FineGym}} (GYM) targets fine-grained action understanding in gymnastics. We use the Gym-99 subset (99 classes) with 20{,}484 training and 8{,}521 test samples, focusing on subtle motion differences within highly structured routines.

\begin{table}[t]
    \centering
    \caption{\textbf{Pretraining configuration.}}
    \label{tab:pretrain_supp}
    \small
    \begin{tabular}{l c c}
        \toprule
        \multicolumn{3}{l}{\textbf{Shared}} \\
        \midrule
        Optimizer & \multicolumn{2}{l}{AdamW} \\
        Base learning rate & \multicolumn{2}{l}{$1.5\times 10^{-4}$} \\
        Weight decay & \multicolumn{2}{l}{0.05} \\
        Momentum (Betas) & \multicolumn{2}{l}{$\beta_1{=}0.9,\ \beta_2{=}0.95$} \\
        Batch size & \multicolumn{2}{l}{512} \\
        LR schedule & \multicolumn{2}{l}{cosine decay} \\
        Warmup epochs & \multicolumn{2}{l}{40} \\
        Augmentation & \multicolumn{2}{l}{MultiScaleCrop{(1, 0.875)}} \\
        \midrule
        \textbf{Dataset-specific} & \textbf{Epochs} & \textbf{FlipAug.} \\
        \midrule
        SSv2  & 800 & no \\
        K400  & 800 & yes \\
        \bottomrule
    \end{tabular}
\end{table}

\begin{table}[t]
    \centering
    \small
    \setlength{\tabcolsep}{3pt}
    \caption{\textbf{Linear probing configuration.}}
    \label{tab:linear_supp}
    \begin{tabular}{l|cccccc}
        \toprule
        config & K400 &  HMDB & SSv2 & GYM  \\
        \midrule
        optimizer & \multicolumn{6}{c}{AdamW} \\
        base learning rate & \multicolumn{6}{c}{1$\times$10$^{-3}$} \\
        weight decay & \multicolumn{6}{c}{0.05} \\
        optimizer momentum & \multicolumn{6}{c}{$\beta_1,\beta_2=0.9,0.999$} \\
        layer-wise lr decay & \multicolumn{6}{c}{0.75} \\
        batch size & \multicolumn{6}{c}{128} \\
        learning rate schedule & \multicolumn{6}{c}{cosine decay} \\
        training epochs & 30  & 100 & 50 & 100  \\
        flip augmentation & \emph{yes} & \emph{yes}  & \emph{no} & \emph{yes} &  \\
        \bottomrule
    \end{tabular}
\end{table}
\begin{table}[t]
    \centering
    \caption{\textbf{Full finetuning configuration.}}
    \label{tab:finetune_supp}
    \small
    \begin{tabular}{l l l l}
        \toprule
        \textbf{Parameter} & \multicolumn{3}{l}{\textbf{Value}} \\
        \midrule
        Optimizer & \multicolumn{3}{l}{AdamW} \\
        Base learning rate & \multicolumn{3}{l}{$1.0\times 10^{-3}$} \\
        Weight decay & \multicolumn{3}{l}{0.05} \\
        Momentum (Betas) & \multicolumn{3}{l}{$\beta_1=0.9,\ \beta_2=0.999$} \\
        Layer-wise LR decay & \multicolumn{3}{l}{0.75} \\
        LR schedule & \multicolumn{3}{l}{cosine decay} \\
        Warmup epochs & \multicolumn{3}{l}{5} \\
        RandAug & \multicolumn{3}{l}{(9, 0.5)} \\
        Label smoothing & \multicolumn{3}{l}{0.1} \\
        Mixup & \multicolumn{3}{l}{0.8} \\
        CutMix & \multicolumn{3}{l}{1.0} \\
        Drop path & \multicolumn{3}{l}{0.1} \\
        \midrule
        \textbf{Dataset-specific} & \textbf{Batchsize} & \textbf{Epochs} & \textbf{FlipAug.} \\
        \midrule
        SSv2   & 32 & 40  & no  \\
        K400   & 16 & 100 & yes \\
        SEVERE & 16 & 100 & yes \\
        \bottomrule
    \end{tabular}
\end{table}

\begin{table}[t]
    \centering
    \caption{\textbf{Datasets details.} Splits of datasets used in  
    full finetuning and linear probing.}
    \label{tab:dataset_details}
    \small
    \setlength{\tabcolsep}{3pt}
    \begin{tabular}{l l c c c}
        \toprule
        \textbf{Dataset} & \textbf{Abbrev.} & \textbf{\#Classes} & \textbf{\#Train} & \textbf{\#Test} \\
        \midrule
        Kinetics\mbox{-}400 & K400 & 400 & 240K & 19K \\
        UCF\mbox{-}101      & UCF  & 101 & 9.5K & 3.8K \\
        HMDB\mbox{-}51      & HMDB & 51  & 4.8K & 2K \\
        Something\mbox{-}Something V2 & SSv2 & 174 & 169K & 24.8K \\
        FineGYM             & GYM  & 99  & 20.5K & 8.5K \\
        \bottomrule
    \end{tabular}
\end{table}
\paragraph{SEVERE Benchmark~\citep{Thoker2022HowSevere}} SEVERE aggregates eight evaluation settings across SSv2, UCF, FineGYM, and Charades to stress sample efficiency, granularity, and task shift. ~\Cref{tab:severe_desc} details each subset and metric.
\begin{table*}[t]
    \centering
    \caption{\textbf{SEVERE benchmark.} Subsets, protocols, and metrics following~\citep{Thoker2022HowSevere}.}
    \label{tab:severe_desc}
    \small
    \setlength{\tabcolsep}{5pt}
    \begin{tabular}{l l l l r r r l}
        \toprule
        \textbf{Dataset} & \textbf{Experiment} & \textbf{Setup Group} & \textbf{Task} & \textbf{\#Classes} & \textbf{Finetune} & \textbf{Test} & \textbf{Metric} \\
        \midrule
        FineGym~\citep{Shao2020FineGym} & Gym99      & Full                 & Action Class.       & 99  & 20{,}484 & 8{,}521 & Top-1 Acc. \\
        \midrule
        UCF 101~\citep{Soomro2012UCF101-arxiv}  & UCF ($10^3$) & Sample Efficiency   & Action Class.       & 101 & 1{,}000  & 3{,}783 & Top-1 Acc. \\
        FineGym~\citep{Shao2020FineGym} & Gym ($10^3$) & Sample Efficiency   & Action Class.       & 99  & 1{,}000  & 8{,}521 & Top-1 Acc. \\
        \midrule
        FineGym~\citep{Shao2020FineGym} & FX\mbox{-}S1 & Action Granularity  & Action Class.       & 11  & 1{,}882  & 777     & Mean-per-class \\
        FineGym~\citep{Shao2020FineGym}  & UB\mbox{-}S1 & Action Granularity  & Action Class.       & 15  & 3{,}511  & 1{,}471 & Mean-per-class \\
        \midrule
        UCFRep~\citep{Zhang2020ContextAware} & UCF\mbox{-}RC & Task Shift       & Repetition Counting & --  & 421     & 105     & Mean Error \\
        Charades~\citep{Sigurdsson2016Hollywood} & Charades & Task Shift       & Multi-label Class.  & 157 & 7{,}985  & 1{,}863 & mAP \\
        \bottomrule
    \end{tabular}
\end{table*}

A summary of the different datasets is presented in \cref{tab:dataset_details}.

\subsection{Training and Evaluation Details}

\paragraph{Pretraining Details.} 
We pretrain on K400~\citep{Kay2017TheKinetics-arxiv} and SSv2~\citep{Goyal2017Something}.  Following VideoMAE~\citep{Tong2022VideoMAE}, we sample clips of 16 frames at $224{\times}224$ with a temporal stride of 2 on SSv2 and 4 on K400.  We compute space–time tube tokens via a 3D convolution, treating each $2{\times}16{\times}16$ cube as a token.  For each sampled 16-frame clip, we temporally downsample it with a stride of 2 to extract the trajectories from CoTracker3~\citep{Karaev2025CoTracker3}, matching the total number of trajectory tokens with space-time cubes.
In practice, we use the normalized temporal differences of the extracted motion trajectories as the target, rather than absolute values.

All hyperparameters are shown in~\cref{tab:pretrain_supp}, following~\citep{Tong2022VideoMAE,Thoker2025SMILE}.  All pretraining runs use 16$\times$NVIDIA A100 GPUs. For downstream tasks, we discard the decoders and use only the pretrained encoder with a task-specific head (\eg, a linear classifier for action recognition).

\paragraph{Linear probing details.}
We strictly follow the settings in~\citep{Thoker2025SMILE} and train the linear head on top of the frozen backbone with the target dataset, as shown in~\cref{tab:linear_supp}. Experiments are run on 4$\times$V100 GPUs.

\paragraph{Full finetuning details}, 
We strictly follow the settings in~\citep{Thoker2025SMILE} and train the backbone + head with the taget dataset, as shown in~\cref{tab:finetune_supp}. Experiments are run on 4$\times$V100 GPUs. 

\paragraph{SEVERE benchmark details.}
Following ~\citep{Thoker2025SMILE,Salehi2024SIGMA}, we evaluate  with the official SEVERE codebase~\citep{Thoker2022HowSevere}, strictly reusing the provided training and evaluation configurations to ensure a fair comparison, also shown in~\cref{tab:finetune_supp}. Experiments are run on 4$\times$V100 GPUs.

\end{document}